# Assessment of algorithms for mitosis detection in breast cancer histopathology images

Mitko Veta[1*], Paul J. van Diest[2], Stefan M. Willems[2], Haibo Wang[3], Anant Madabhushi[3], Angel Cruz-Roa[4], Fabio Gonzalez[4], Anders B. L. Larsen[5], Jacob S. Vestergaard[5], Anders B. Dahl[5], Dan C. Cireşan[6], Jürgen Schmidhuber[6], Alessandro Giusti[6], Luca M. Gambardella[6], F. Boray Tek[7], Thomas Walter[8,9,10], Ching-Wei Wang[11], Satoshi Kondo[12], Bogdan J. Matuszewski[13], Frederic Precioso[14], Violet Snell[15], Josef Kittler[15], Teofilo E. de Campos[15, 16], Adnan M. Khan[17], Nasir M. Rajpoot[17,18], Evdokia Arkoumani[19], Miangela M. Lacle[2], Max A. Viergever[1], Josien P.W. Pluim[1]

[1] Image Sciences Institute, University Medical Center Utrecht, Utrecht, The Netherlands
[2] Pathology Department, University Medical Center Utrecht, Utrecht, The Netherlands
[3] BME Department, Case Western Reserve University, Cleveland, USA
[4] MindLab, National University of Colombia, Bogota, Colombia
[5] Department of Applied Mathematics and Computer Science, Technical University of Denmark, Kongens Lyngby, Denmark
[6] IDSIA, USI-SUPSI, Lugano, Switzerland
[7] Department of Computer Engineering, Işık University, İstanbul, Turkey
[8] MINES ParisTech, PSL-Research University, CBIO-Centre for Computational Biology, France
[9] Institut Curie, Paris, France
[10] INSERM U900, Paris, France
[11] Graduate Institute of Biomedical Engineering, National Taiwan University of Science and Technology, Taipei, Taiwan
[12] Corporate R&D Center, Panasonic Healthcare Co., Ltd., Osaka, Japan, currently with: Konica Minolta, Inc.
[13] School of Computing, Engineering and Physical Sciences, University of Central Lancashire, Preston, UK
[14] I3S, UMR 7271 UNS-CNRS, Nice Sophia-Antipolis University, Sophia Antipolis, France
[15] CVSSP, University of Surrey, Guildford, UK
[16] Department of Computer Science and Sheffield institute for Translational Medicine, University of Sheffield, UK
[17] Department of Computer Science, University of Warwick, UK
[18] Department of Computer Science & Engineering, College of Engineering Qatar University, Doha, Qatar
[19] The Michael Letcher Department of Cellular Pathology, Princess Alexandra Hospital NHS Trust, Harlow, Essex, UK

* Corresponding author: e-mail: mitko@isi.uu.nl, phone: +31 88 75 58353, Image Sciences Institute, University Medical Center Utrecht, Heidelberglaan 100, 3584 CX Utrecht, The Netherlands

# Abstract

The proliferative activity of breast tumors, which is routinely estimated by counting of mitotic figures in hematoxylin and eosin stained histology sections, is considered to be one of the most important prognostic markers. However, mitosis counting is laborious, subjective and may suffer from low inter-observer agreement. With the wider acceptance of whole slide images in pathology labs, automatic image analysis has been proposed as a potential solution for these issues.

In this paper, the results from the Assessment of Mitosis Detection Algorithms 2013 (AMIDA13) challenge are described. The challenge was based on a data set consisting of 12 training and 11 testing subjects, with more than one thousand annotated mitotic figures by multiple observers. Short descriptions and results from the evaluation of eleven methods are presented. The top performing method has an error rate that is comparable to the inter-observer agreement among pathologists.

**Keywords:** Breast cancer, whole slide imaging, digital pathology, mitosis detection, cancer grading.

# 1   Introduction

Breast cancer patients can considerably benefit from adjuvant therapy. However, aggressive adjuvant therapies are costly, can lead to potentially serious side effects and thus are only given to patients that are at a high risk. Assessing the patient risk requires use of good prognostic factors. In this regard, prognostic factors related to tumor proliferation have proven to be among the most powerful ones (Diest et al., 2004).

The proliferation of cells occurs through a process that can be divided into several phases: resting phase (G0), first gap phase (G1), synthesis phase (S), second gap phase (G2) and mitotic phase (M). After the M-phase, the cells either enter the G0-phase or the G1-phase repeating the process. The cells that are in the M-phase can be visually determined under a microscope by their characteristic morphology. In hematoxylin and eosin (H&E) stained breast cancer sections, mitoses are discernible as hyperchromatic objects that lack a clear nuclear membrane and have therefore specific shape properties. Counting of mitotic figures in H&E stained sections is the oldest and still most widely used form of assessment of proliferation of breast cancer tumors by pathologists.

The proliferative activity of the tumor is estimated as the number of mitoses in an area of 2 mm$^2$, which corresponds to 8-10 microscope high power fields (HPFs; refers to the area that is visible using the microscope under very high magnification, usually ×40) depending on the microscope model. This number is referred to as the mitotic activity index (MAI). Mitosis counting is routinely performed in pathology labs all over the world and is widely used as a prognostic factor. Although MAI assessment can be well reproducible if a strict protocol is followed after rigorous training (van Diest et al., 1992), it is a subjective procedure that is liable to intra-observer variation. Several factors contribute to this. First of all, the task of identifying mitotic figures in H&E sections is not trivial. They can display a number of different appearances, with their hyperchromacity being the most salient feature. Moreover, many other cellular components can have a similar hyperchromatic appearance, such as apoptotic or necrotic nuclei, compressed nuclei, "junk" particles and other artifacts from the tissue preparation process. This makes the identification of mitoses difficult. Furthermore, the assessment of the proliferative state by counting mitotic figures is performed only in a small area of the tumor selected to be at the tumor periphery and to have high cellularity. The choice of the area is also a matter of subjective interpretation and one of the potential sources of low reproducibility.

In addition to being subjective, mitosis counting is a laborious task, compared with the assessment of other prognostic factors for breast tumors, such as nuclear pleomorphism and tubule formation. For a typical case, it takes 5-10 min. for a pathologist to perform mitosis counting, and the process must sometimes be repeated in different areas or different tumor slides for borderline cases.

In the last decade, pathology labs have started to move towards a fully digital workflow, with the use of digital slides being the main component of this process (Stathonikos et al., 2013). This was made possible by the introduction of scanners for whole slide imaging (WSI) that enable cost-effective production of digital representations of glass slides. In addition to many benefits in terms of storage and browsing capacities of the image data, one of the advantages of digital slides is that they enable the use of image analysis techniques that aim to produce quantitative features to help pathologists in their work. An automatic mitosis detection method with good performance could alleviate both the subjectivity and the tediousness of manual mitosis counting, for example, by independently producing a mitotic activity score or guiding the pathologist to the region within the tissue with highest mitotic activity.

This paper gives an overview of the Assessment of Mitosis Detection Algorithms 2013 (AMIDA13) challenge that we recently launched. The main goal of the challenge was to evaluate and compare the performance of different (semi-)automatic mitosis detection methods that work on regions extracted from whole slide images on a large common data set. Since only the number of mitoses present in the tissue is of importance, i.e. their size and shape is not of interest, the challenge was defined as a detection problem.

## 1.1 Challenge format

The challenge was opened on March 28th, 2013, at which point interested groups or individuals could register on the challenge website[1] and download the training data set that they could use to develop their methods. The training data set consisted of image data accompanied by ground truth mitosis annotations. Approximately two months after the release of the training data set a testing data set of similar size was released. The testing data set consisted only of image data, i.e. the ground truth annotations were withheld by the challenge organizers in order to ensure independent evaluation. After the release of the testing data set, the participants were able to run their methods and upload results to the challenge website for evaluation. The number of submissions from each registered participant was limited to three in order to avoid overfitting of the method to the testing data. Each submission had to be accompanied by a short method description, or, in the case it was the second or third submission of the participating team, a description of how the method differed from previous submissions. The submitted results were evaluated by the challenge organizers and the evaluation result was made available to the participants.

The first part of the AMIDA13 challenge was concluded with the workshop held in conjunction with the Medical Image Computing and Computer Assisted Interventions (MICCAI) 2013 conference on September 22nd, 2013 in Nagoya, Japan. The deadline for submissions that were to be presented at the challenge workshop and included in this overview paper was September 8th, 2013. After the conclusion of the workshop, the challenge website was reopened for new submissions.

Prior to the challenge workshop, over 110 teams or individuals from more than 30 countries registered to download the data set. Fourteen teams submitted results for evaluation. All submissions were automatic methods. Although the option was provided, none of the submissions was a semi-automatic method. This overview paper includes eight methods proposed by teams that submitted results for evaluation and attended the challenge workshop. Additionally, three methods by teams that submitted results for evaluation and achieved good performance but could not attend the challenge workshop are also included. The full list of results along with any new submissions after the challenge workshop is available on the challenge website.

## 1.2 Previous work

The earliest methods for automatic mitosis detection in breast cancer tissue date back to more than two decades ago (Beliën et al., 1997; Kaman et al., 1984; Ten Kate et al., 1993). However, those approaches were inevitably constrained in their performance and potential use by the limited slide digitization technology and available computational resources at that time. The recent interest in this problem (Ciresan et al., 2013; Malon et al., 2012; Veta et al., 2013) was ignited by the increased availability of WSI scanners. This resulted in the organization of the MITOS challenge (Roux et al., 2013) in 2012 on the same topic, which was well attended and helped advance the state of the art for this problem. However, the challenge was based on a data set of relatively small size (5 slides in total, 10 annotated HPFs per slide) and it did not account for the inter-subject variability in tissue appearance and staining as regions of the same slides were included in both the training and testing data sets. We aim to address these issues with the data set used in this challenge.

For a wider overview of histopathology image analysis techniques for breast cancer and other tissue types, we refer the reader to the recent reviews in Gurcan et al. (2009) and Veta et al.

---

[1] http://amida13.isi.uu.nl

(2014). We would also like to point out that automatic mitosis detection is a relevant application for other microscopy imaging modalities such as live cell images (Harder et al., 2009; Padfield et al., 2011) and there is a possibility for cross-over of the proposed methods.

## 2 Materials

In this section, we describe the process by which the challenge data set was compiled and annotated.

### 2.1 Patient, slide and region selection

The histology slides that were used for the creation of the challenge data set were produced at the Pathology Department of the University Medical Center Utrecht, Utrecht, The Netherlands. This is a pathology laboratory of medium size handling more than 144,000 surgical pathology slides and 12,000 cytology slides each year. From the archives of the department, slides from 23 consecutive invasive breast carcinoma patients, admitted between July 2009 and January 2010, were retrieved based solely on their availability (other selection criteria were not employed). All slides were prepared according to the standard laboratory protocol that consists of formalin fixation and paraffin embedding of the tissue, followed by cutting of 3-5 µm thick sections and staining with H&E.

One of the most difficult challenges in histopathology image analysis is the variability of tissue appearance, which is mostly the result of the variability in the conditions of the tissue preparation and staining processes. The challenge data set consists of slides that were routinely prepared at different time points during a longer period of time. In this way, it can be expected that appearance variability, which can be avoided when the tissues are processed in a single batch, will be reflected in the data set and the challenge will thus provide realistic performance estimates.

Data of a single patient typically consist of multiple slides. In clinical practice, the pathologist selects the slide and the region within the slide that is most suitable for the analysis at hand. Automating these selection steps is interesting by itself, and is certainly needed for a fully automatic workflow, however, for the challenge data set we decided to perform these steps manually and focus the challenge on the problem of mitosis detection.

An expert pathologist selected one representative slide per patient and marked a large region of the tumor on the glass slides in which mitosis annotation was to be performed. For the larger tumors, the marked areas within the slides were selected to encompass the most invasive part of the tumor, to be located at the periphery and to have high cellularity, which are the standard guidelines for performing mitosis counting. Smaller tumors were included in their entirety. The size of the outlined area varied from 7 $mm^2$ to 58 $mm^2$ with a median of 26 $mm^2$. It should be noted here that this diverges from the routine pathology practice of estimating the mitotic activity as the number of mitotic figures within an area of 2 $mm^2$. The choice of marking a larger area for mitosis annotation was made in order to ensure that a larger number of mitoses could be identified, which would result in data set of a size that is sufficient for training and evaluation of an automatic mitosis detection method.

### 2.2 Image acquisition

The representative regions were digitized with a ScanScope XT whole slide scanner (Aperio, Vista, CA, USA). This scanner model can perform the steps of tissue selection, patch focus point selection, calibration, image acquisition and compression in a fully automatic manner. The digitization of the candidate regions was done at 40× magnification with a spatial resolution of 0.25 µm/pixel. The automatic focus points were manually reviewed before single focus plane scanning to reduce the possibility of blurry patches. High quality JPEG2000 compression (quality factor 85) was used to store the images, almost completely eliminating any visible compression artifacts, which was confirmed with side-by-side comparison of compressed and uncompressed versions of several regions. The slide digitization parameters were optimized to ensure the highest image quality possible and differs from the standard practice of digital slide archiving at the UMC Utrecht (Huisman et al., 2010).

**Table 1** – Data set summary.

| ID | Outlined area (mm$^2$) | Number of annotated objects by observer 1 | Number of annotated objects by observer 2 | Number of HPFs in the challenge data set | Number of ground truth mitotic figures |
|---|---|---|---|---|---|
| **Training data set** | | | | | |
| 1 | 20 | 60 | 188 | 39 | 73 |
| 2 | 36 | 51 | 32 | 28 | 37 |
| 3 | 41 | 13 | 46 | 16 | 18 |
| 4 | 24 | 178 | 338 | 61 | 224 |
| 5 | 20 | 5 | 8 | 10 | 6 |
| 6 | 43 | 104 | 89 | 61 | 96 |
| 7 | 22 | 55 | 140 | 43 | 68 |
| 8 | 21 | 2 | 5 | 10 | 3 |
| 9 | 58 | 0 | 12 | 10 | 2 |
| 10 | 26 | 0 | 4 | 10 | 0 |
| 11 | 20 | 5 | 25 | 13 | 15 |
| 12 | 41 | 9 | 5 | 10 | 8 |
| **Testing data set** | | | | | |
| 1 | 7 | 2 | 4 | 10 | 3 |
| 2 | 36 | 1 | 37 | 15 | 16 |
| 3 | 40 | 69 | 73 | 49 | 66 |
| 4 | 21 | 11 | 10 | 10 | 9 |
| 5 | 27 | 8 | 9 | 10 | 6 |
| 6 | 26 | 277 | 247 | 67 | 212 |
| 7 | 7 | 0 | 2 | 10 | 2 |
| 8 | 26 | 0 | 5 | 10 | 0 |
| 9 | 31 | 111 | 155 | 44 | 115 |
| 10 | 34 | 93 | 133 | 48 | 72 |
| 11 | 23 | 34 | 32 | 22 | 32 |

## 2.3 Ground truth annotation

The ground truth for the challenge was assigned based on the annotations by multiple observers, to reduce the influence of observer variability. We used the following protocol to establish the ground truth mitosis annotations.

1. Two pathologists independently traversed the indicated regions of interest and marked the locations of mitoses by drawing an ellipse encompassing the object in the whole slide image viewer;

2. The objects on which both pathologists agreed were directly taken as ground truth mitosis objects. Two annotations were considered to indicate the same mitotic figure if the Euclidian distance between their centers was less than 7.5 μm (30 pixels), which is the same criterion used to evaluate the automatic detection methods;

3. The discordant objects (annotated as mitoses by only one of the observers) were presented to a panel of two additional pathologists to make the final decision. Note that the panel did not traverse the slides but only examined the discordant objects.

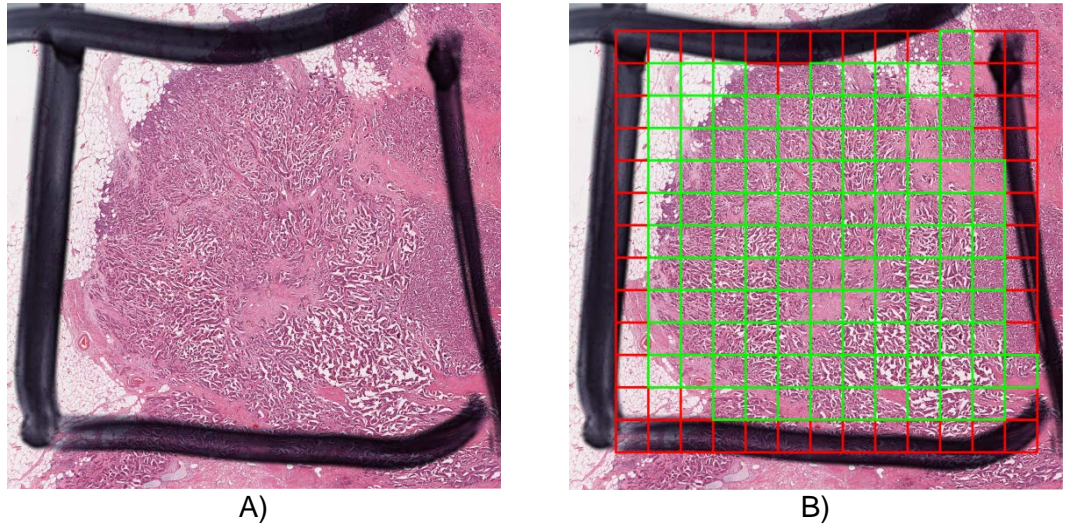

**Figure 1** – Separation into high power fields (HPFs). A) Example slide with the area for annotation indicated with a black marker. B) Each rectangle from the grid is one HPF. HPFs that intersect the black marker lines (given in red) are not included in the data set.

With this annotation protocol, all objects that were accepted as ground truth mitoses have been agreed upon by at least two experts.

The first set of annotations was done by a pathologist at the UMC Utrecht using the ImageScope whole slide viewer (Aperio, Vista, CA, USA). The second set of annotations was done by an external pathologist using pathoconsult.nl – an online digital slide viewing and collaboration platform maintained by the UMC Utrecht. The second observer was blinded to the results from the first observer. The observers did not receive a standardized definition for mitotic figures, but were instructed to perform the mitosis counting using the criteria they employ in daily practice.

The total number of annotations made by the first and second observers was 1088 and 1599, respectively. The number of locations upon which they agreed was 649, which left 1389 annotations to be resolved by the panel of two additional observers. The panel revisited all discordant objects and after discussion, decided together which objects to accept as ground truth mitoses. The total number of remaining objects after the consensus annotation was 1157 (this number also includes the concordant objects that were directly accepted as ground truth).

We note here that although the difference in the absolute counts between the two initial observers is quite large (1088 compared to 1599), a large portion of it can be traced back to only a few subjects (Table 1). Later investigation into this showed that this discrepancy can be largely attributed to the difference in the interpretation of objects that are difficult to interpret without fine focusing ability. It is also reasonable to assume that some of the difference can be attributed to the inter-institutional differences in mitosis counting.

## 2.4 Image data distribution

For simplicity, the image data for the challenge was not distributed as whole slide images, but instead, each whole slide image was divided into a number of smaller TIFF images that could be easily read by most image analysis platforms. We chose the size of the TIFF images to be 2000×2000 pixels, representing an area of 0.25 mm$^2$, which is in the order of one microscope high power field (the exact area of one HPF is different for different microscope models). We refer to the TIFF images as HPFs. HPFs that were outside the marked region of interest on the slide or that intersected the black marker annotation that indicates the region of interest were excluded (Figure 1B). Since for some subjects the total number of HPFs was very high (in the order of several hundreds), only the HPFs that contained at least one mitosis were included as part of the data set. For the subjects that had fewer than 10 HPFs with mitosis occurrence, additional empty HPFs were included to extend the total number to 10 in order to include sufficient background image data, necessary for good training and evaluation. Note that some mitotic figures fell into a HPF that intersects the black marker annotation and are thus not included in the data set. This was the case with around 6% of the annotations.

The set of 23 subjects was split into two subsets – one intended for training of the methods and one used as an independent testing data set. The division into training and testing data sets was done in such a way that the number of HPFs and ground truth mitoses was approximately balanced. A summary of the two data sets is given in Table 1. The ground truth was provided to the participants only for the training data set, in the form of coordinates of mitoses (the centers of the elliptical annotations) for each HPF.

## 2.5 Object-level observer agreement

Given the notorious difficulty of the mitosis detection task even for expert pathologists, the performance of the automatic mitosis detection methods should be evaluated within the context of the inter-observer agreement.

The similarity of two sets of annotations can be expressed using the Dice overlap coefficients, which is computed as:

$$D(A, B) = \frac{2|A \cap B|}{|A| + |B|}$$

where *A* and *B* are the two sets of annotations and $|\cdot|$ indicates the number of elements in the sets.

On the entire data set of 23 subjects, prior to the separation and selection of HPFs, the Dice overlap coefficient was 0.483 (i.e., there were 2687 annotations by the two observers in total and they agreed for 649 objects). For the HPFs in the training data set, the Dice overlap coefficient between the sets of annotations by the two independent observers was 0.527, and for the testing data set 0.566. These numbers are higher[2] compared to the value for the entire data set (prior to separation into HPFs and rejection) due to the selection of non-empty HPFs based on the consensus annotation.

The Dice overlap coefficients between the individual annotations and the consensus annotation are 0.749 and 0.763 for the first and second observer respectively for the selected HPFs in the training data set, and 0.796 and 0.773 for the selected HPFs in the testing data set. It should be noted however that this is a biased measure since the consensus annotation is partly based on the two independent annotations.

## 3 Methods

### 3.1 CCIPD/MINDLAB[3]

**Preprocessing**: In addition to the three color channels from the RGB input images, four additional color channels were computed and used for candidate segmentation and feature extraction: L from the LAB color space, V and L from the LUV color space and the blue ratio image (ratio of the blue color channel and the sum of the other two channels in the RGB color space).

---

[2] Note that the annotation procedure led to rejection of some HPFs that had only discordant annotations by the two initial observers that were then in turn all rejected by the consensus annotation. This produces higher values for the Dice overlap when only the selected HPFs are considered.
[3] CCIPD at Case Western Reserve University, USA and MindLab at National University of Colombia

**Candidate detection and segmentation**: For each HPF, a set of candidate regions was defined by thresholding the blue ratio image.

**Feature extraction and classification**: This method fuses two classification strategies: a feature learning method based on Convolutional Neural Networks (CNN), and a set of handcrafted features combined with a random forests (RF) classifier (Wang et al. 2014). For each candidate region, both learned features and handcrafted features were extracted independently, and then classified using the corresponding classifier.

The CNN model has 4-layer architecture, including two consecutive convolution-pooling layers, a fully-connected layer and a softmax classification layer. It operates on 80×80 pixel patches in the YUV color space centered at the candidate regions. The first 3 layers comprise 64, 128, and 256 neurons, respectively. For the convolution-pooling layers, fixed 8×8 convolutional kernel and 2×2 pooling kernel were applied. The CNN was trained using stochastic gradient descent.

In addition, morphological, statistical and texture features were extracted for each of the seven color channels. Principal component analysis (PCA) was applied to reduce the dimension of the extracted features by retaining 98.5% of the principal components. Using this reduced representation, a cascade of two RF classifiers with 100 random trees was trained. To balance the numbers of mitosis and non-mitosis objects, the number of non-mitosis nuclei was reduced by eliminating overlapping objects and oversampling the positive class.

The final prediction score is a weighted average of the outputs of the two classifiers. More details about this method can be found in Wang et al. (2014).

## 3.2 DTU[4]

**Preprocessing:** Candidate detection was performed on the blue ratio image, calculated as the ratio of the blue color channel and the sum of the red and green channels.

**Candidate detection and segmentation**: Candidate detection was performed by thresholding of the Gaussian of Laplacian blob detector applied to the blue ratio image, followed by connected component labelling.

**Feature extraction and classification**: For each detected candidate object, a 100×100 pixel patch was extracted and each RGB color channel was independently normalized to have values in the range [0 1]. From the color normalized patches, three types of image features were extracted:

1. Color intensity histograms, one for each color channel;
2. Gradient orientation histograms. The orientations of the gradients are calculated relative to a vector from the cell center to the location of the gradient;
3. Shape index histograms (Larsen et al., 2014). The shape index captures second-order differential structure from the local Hessian eigenvalues. The two eigenvalues are mapped to a continuous interval providing a smooth and intuitive transition between the second-order shapes (cup, rut, saddle, ridge and cap).

Each image feature was computed for different concentric donut-like spatial pooling regions centered on the candidate object. The donut-shaped pooling regions vary in radius and width such that they capture image structure in different parts of the candidate object. Note that the features are rotationally invariant because both the image features and the spatial decomposition are rotationally invariant.

The extracted image features were used to train a support vector machine (SVM) classifier with radial basis function (RBF) kernel.

---
[4] Technical University of Denmark

## 3.3 IDSIA[5]

In this approach, Multi Column Max-Pooling Convolutional Neural Networks (MCMPCNN) are used for supervised pixel classification. MPCNNs alternate convolutional layers with max-pooling layers. A similar technique won the MITOS mitosis detection competition (Ciresan et al., 2013) and recently produced outstanding results in image classification (Ciresan et al., 2012b) and segmentation (Ciresan et al., 2012a). The inputs to the MPCNN are 63×63 pixel patches directly sampled from the raw RGB images. The output is the probability that the central pixel of the patch is within 20 pixels of the centroid of a mitosis. Three networks with identical 10-layer architecture were trained on 20 million patches extracted from the training images. One tenth of such instances were randomly sampled from mitosis pixels (which represent a tiny fraction of all pixels in the training data set); 40% were randomly sampled from all non-mitosis pixels; the remaining 50% were sampled only from non-mitosis pixels that were found to be similar to mitosis (therefore, more challenging to classify) during a simple preprocessing phase. The resulting training data set was augmented by adding rotated and mirrored instances. Each network was trained for a maximum of five epochs, which required about three days of computation using an optimized GPU implementation. Mitoses in the test images were detected by sliding the neural network detectors over the images by means of an efficient algorithm (Giusti et al., 2013). This resulted in a map where each pixel represented the probability of belonging to the mitosis class. Each test image was processed in eight different rotation/mirroring combinations by each of the three networks. The 24 resulting probability maps were averaged, and then convolved with a 20-pixel radius disk kernel. Nonzero values obtained after performing non-maxima suppression in a 50-pixel radius corresponded to detected centroids of mitotic figures.

## 3.4 ISIK[6]

This method is an extension of previous work on mitosis detection (Tek, 2013).

**Preprocessing**: Prior to candidate detection, image contrast stretching was performed.

**Candidate detection and segmentation**: Candidate objects were initially segmented by a morphological double threshold operation. The resulting binary image was filtered by an area opening operation (with the minimum area set to 50 pixels), to remove isolated spurious regions. The candidate extraction procedure was finalized with a morphological hole filling step.

**Feature extraction and classification**: The candidate object classification consists of two separate stages. In the first stage, a set of simple object features were employed to significantly reduce the number of false objects, while keeping the loss of true mitotic figures to a minimum. The following set of features was used in this stage: area, major- and minor-axis lengths, perimeter, equivalent diameter, ratio of the area to the perimeter, eccentricity, extent, mean intensities of the three RGB channels and the ratios of the mean intensities of the three RGB channels to the corresponding means of all candidate components of the same image.

In the second stage, which operates on candidate objects that have not been eliminated in the first stage, a set of windows of increasing width around the candidate object were defined. For each window, a feature vector that is formed of five different groups of features is calculated: color (statistics of the RGB color channels and similarities to the average mitosis and non-mitosis histograms), binary shape, Laplacian, morphological (area granulometry) and gray-level co-occurrence. For both stages of the classification, an ensemble of multi-stage AdaBoost classifiers was used.

---

[5] IDSIA, Dalle Molle Institute for Artificial Intelligence, USI-SUPSI, Lugano, Switzerland
[6] Department of Computer Science and Engineering, Isik University, İstanbul, Turkey

## 3.5 MINES[7]

**Preprocessing**: Separate hematoxylin and eosin channels were obtained with color unmixing (deconvolution) (Ruifrok and Johnston, 2001).

**Candidate detection and segmentation**: The segmentation of candidate objects was performed entirely using the hematoxylin channel. In order to detect potential nuclei (candidates), a diameter closing operation (Walter et al., 2007) was applied to the median filtered image removing all dark structures with maximal extension smaller than a predefined parameter (the diameter was chosen to be 80 pixels). By calculating the difference to the median filtered image, these small dark structures could be segmented by simple double thresholding (low threshold: 34, high threshold: 60). The connected components of this binary image were considered candidate objects.

**Feature extraction and classification**: The candidate segmentation procedure failed to identify only 2 mitoses in the training data set. With the aim to reduce the high number of false positives, a supervised classifier was trained. For each candidate object, shape and texture features (basic geometric and gray level features, Haralick features, statistical geometric features, morphological granulometries, convex hull features, etc.), as defined in Walter et al. (2010), were computed. In addition, Haralick and basic gray level features were calculated for the candidate region in the eosin channel and for a ring around each candidate region in the hematoxylin channel in order to quantify the local environment.

A training data set of three classes was built: non-mitosis, early mitosis (prophase/prometaphase) and late mitosis (metaphase, anaphase). The rationale of transforming the binary classification problem into a three-class problem was that the two mitosis classes are morphologically very different and some preliminary runs showed that this strategy gives better results (for this particular problem). In order to distinguish the three classes, an SVM classifier was trained (RBF kernel, parameters automatically determined by a grid search with 5-fold cross validation). The mitotic cells were taken as the union of the early mitosis and the late mitosis class obtained from this classifier.

## 3.6 NTUST[8]

In this approach, a diverse cascade learning framework (Wang and Hunter, 2010a) with the cwBoost learning algorithm (Wang and Hunter, 2010b) is used for supervised pixel classification. The hierarchical ensemble classifier contains 10 layers of simple mitosis detectors, which evaluate various types of inputs with different models, and can quickly filter out negative areas. A similar technique was used for obscured human head detection in video sequences (Wang and Hunter, 2010a). The inputs to the learning methods were 50×50 pixel patches directly sampled from image data extracted from the red color channel. The red color channel was chosen because in preliminary tests it was found to give the best cross-validation accuracy compared to the other color channels. Ten layers of boosting ensembles were trained on image patches extracted from the training images. Each ensemble contained 10 C4.5 decision tree classifiers (Quinlan, 1996). Mitoses in the test images were then detected by sliding the hierarchical boosting detectors over the images. The output from the detector was the probability that the central pixel of the patch is within 25 pixels of the centroid of a mitosis. A confidence weight is generated by computing the number of detections in the same area, and regions with a weight greater than 2 are defined as possible mitosis regions.

## 3.7 PANASONIC[9]

**Preprocessing**: The RGB images were first transformed into a number of different color spaces that later facilitated the candidate segmentation and feature extraction: L*a*b, HSV, blue ratio image (BR), red ratio image (RR) and blue-red ratio image (BRR). The BR image, which accentuates the nuclear dye, was computed as the ratio of the blue channel and the sum of the other two channels. The RR and BRR images were computed in a similar manner.

---

[7] Centre for Computational Biology - Mines ParisTech, Institut Curie and U900 INSERM, Paris, France
[8] Graduate Institute of Biomedical Engineering, National Taiwan University of Science and Technology
[9] Panasonic Healthcare Co., Ltd., Osaka, Japan

**Candidate detection and segmentation**: Candidate mitosis regions were extracted by binary thresholding of the BR image. The threshold was automatically determined as three times the standard deviation of the BR image. Regions that were smaller than 80 pixels were eliminated.

**Feature extraction and classification**: The following morphological features were computed for each candidate region: area, major axis length, minor axis length, eccentricity, orientation, convex area, filled area, equivalent diameter, solidity, extent, and perimeter. In addition to this, for each candidate region, a rectangular window was defined, and the following features were extracted within the window:

1. Histogram of local binary patterns (LBP) for the BR, RR, BRR, L*, H, S, and V images. The LBP features were computed for three radii (1, 3 and 5 pixels), and the histograms were concatenated;
2. Haralik features (contrast, correlation, energy, and homogeneity) for the BR, RR, BRR, L*, H, S, and V images;

The candidate objects were classified as mitoses or non-mitoses using a random forest classifier.

### 3.8  POLYTECH/UCLAN[10]

**Preprocessing**: The candidate detection was performed in the blue corrected images, which were intensity adjusted to calibrate both the image contrast and the average intensity, partially compensating for the differences in tissue appearance.

**Candidate detection and segmentation**: Candidate objects were detected by thresholding and binary morphological operations. Patches of 128×128 pixel were then extracted around the centroid of each candidate object and subsequently used for feature extraction and classification.

**Feature extraction and classification**: The features used for classification of the candidate objects were selected to represent both textural and shape information. The first set of 10 features was extracted from the average of the run–length matrices calculated in four directions {0°,45°,90°,135°} (Irshad, 2013). The second set of eight features was extracted from the average of the co-occurrence matrices for the same four directions and includes: energy, entropy, correlation, difference moment, inertia, cluster shade, cluster prominence and Haralick's correlation. To capture spatial information, each patch was divided into seven rings and a central circle, and for each region an eight-bin intensity histogram was calculated giving in total 64 features. The final feature that was calculated is the area of the segmented candidate region by binary thresholding. The OpenCV implementation of random forests was used for classification. Due to the highly imbalanced training data set, additional positive (mitosis) patches were randomly selected within a small neighborhood of the ground truth mitosis location.

### 3.9  SURREY[11]

**Preprocessing**: To compensate for the variability of the tissue staining and preparation, the images were first aligned in color space. This was done using histogram matching, with the mean histogram from the whole training data set as the target, and the histograms of the individual subjects as the input. Each color channel was adjusted independently. The computed histograms excluded pixels that belong to regions that do not contain tissue (i.e., white regions), found by thresholding of the green channel.

---

[10] University Nice - Sophia Antipolis, France and University of Central Lancashire, UK
[11] Centre for Vision, Speech and Signal Processing (CVSSP), University of Surrey, UK

**Candidate detection and segmentation**: Candidate mitosis locations were detected based on color. Each color channel was quantized to 64 levels, and these values were used to address a 3-dimensional (RGB) lookup table that points to the likelihoods of the color being present in a mitotic figure. The color lookup table was defined based on the histograms of 10-pixel circular neighborhoods of ground truth locations.

After obtaining a likelihood map for an input image, it was low-pass filtered and thresholded, and the centers of the connected component regions were taken as candidate locations. Around each candidate location, 70x70 pixel patches were extracted and converted to grayscale. Up to two largest objects within the patch were segmented by a threshold that provides the best combination of high boundary gradient and low variance within the object(s). Objects that had area and contrast with the background below predefined thresholds were removed. When a pair of objects was segmented, it was ensured that they had roughly the same area and intensity.

**Feature extraction and classification**: For each candidate, a set of rotation invariant features reflecting the shape, contrast, edge properties and texture of both the segmented object and the background was calculated. In addition, pairs were characterized by the ratio and average of a subset of parameters from each of the objects. For more details on the descriptors, please refer to the Results page on the challenge website. For classification, RBF SVM classifier was used with dominant class subsampling and model averaging, to deal with the class imbalance.

## 3.10 SHEFFIELD/SURREY[12]

This method requires a minimal input in its design and models the space of mitosis images using a low-dimensional manifold. An advantage of manifold learning models is that they enable practitioners to easily visualize the range of mitosis appearances.

The preprocessing and candidate extraction steps of this submission are the same as the ones described in SURREY, but instead of computing a set of predefined features, the normalized gray-level candidate patches were represented as vectors and modeled as observations by a Bayesian Gaussian Process Latent Variable Model (BGPLVM) (Titsias and Lawrence, 2010). This method learns a low dimensional latent space that is mapped nonlinearly back to the original space of observations. In addition, it enables the computation of an approximate density function of novel samples given the known image samples. Therefore, one BGPLVM was used for the positive samples (mitotic cells) and another for the negative samples (false-positive candidate objects). The two classes were assumed to be independent and classification was done using maximum likelihood. Visual inspection of the models (by reconstructing samples in different positions of the latent space) showed that images reconstructed from the positive model were sharper than those of the negative samples, i.e, there was a higher appearance variation among negative samples. Furthermore, by navigating through the positive latent space, the reconstructed images showed a smooth transition between different phases of mitosis as well as between different types of mitosis appearances (including tripolar mitoses that occur in cancer cells). However, the classification results were relatively poor, probably because the assumption of independence between the classes does not hold, indicating that a method that jointly models samples and labels can be more promising (e.g. using Manifold Relevance Determination (Damianou et al., 2012)).

## 3.11 WARWICK[13]

**Preprocessing:** Staining normalization by non-linear color mapping (Khan et al., 2014) was performed in order to neutralize the inherent variation in the color of the staining.

---

[12] Department of Computer Science and Sheffield Institute for Translational Medicine, University of Sheffield, UK and Centre for Vision, Speech and Signal Processing (CVSSP), University of Surrey, UK
[13] University of Warwick, UK

**Candidate detection and segmentation:** Candidate objects were extracted by statistical modelling of the pixel intensities of the b-channel from the Lab color space. The pixel intensities from mitosis regions were modeled by a Gamma distribution and those from non-mitosis regions by a Gaussian distribution (Khan et al., 2013). Maximum-likelihood estimation was employed in order to estimate the unknown parameters of the distributions. The posterior probability map was then binarized to identify candidate objects. The threshold value for obtaining the binary map was selected by performing receiver operating characteristic (ROC) curve analysis. All candidate objects with an area between 40 and 500 pixels were considered candidate objects. This range of areas was obtained by examining the size of the ground truth mitoses in the training data set.

**Feature extraction and classification:** For each candidate object, a set of object level color, shape and texture features was computed. In addition, a small context window around the candidate objects was defined, and used to compute contextual features (first order statistics over a set of textural features). These contextual features were combined with the object features to train a classifier.

Since the number of non-mitosis candidate objects was disproportionately higher compared to mitosis candidate objects, the classification was performed with the RUSBoost classifier (Seiffert et al., 2010), which combines under-sampling and boosting to handle the class imbalance problem. Other classifiers (such as SVM, LDA) were also evaluated, but their cross-validation performance on training data proved to be lower than RUSBoost.

## 4 Evaluation

A detected object was considered to be a true positive if the Eucledian distance to a ground truth location is less than 7.5 μm (30 pixels). This value corresponds approximately to the average size of mitotic figures in the data set, and provides a reasonable tolerance for misalignment of the ground truth location and the detection. When multiple detections fell within 7.5 μm of a single ground truth location (eg., when two components of a single mitotic figure were detected separately), they were counted as one true positive. All detections that were not within 7.5 μm of a ground truth location are counted as false positives. All ground truth locations that do not have detected objects within 7.5 μm were counted as false negatives.

For each proposed method two types of evaluation measures relating to the detection accuracy were computed. The first evaluation measure was the overall $F_1$-score, where all ground truth objects were considered a single data set regardless to which patient they belong. The $F_1$-score[14] is defined as the harmonic mean of the precision (*Pr*; positive predictive value) and the recall (*Re*; sensitivity):

$$F_1 = \frac{2Pr \times Re}{Pr + Re}; Pr = \frac{TP}{TP + FP}; Re = \frac{TP}{TP + FN},$$

where TP, FP and FN are the number of true positive, false positive and false negative detections, respectively. The overall $F_1$-score is dominated by the subjects that have a high number of mitotic figures. To complement this measure, individual $F_1$-scores for each subject were also calculated. For the submissions that included probability estimates associated with the detections, the precision-recall (PR) curves were also computed.

After a visual inspection of the detection results, it was observed that many of the false positives produced by the top performing methods closely resemble mitotic figures. Indeed, owing to the difficulty of the task it is possible that some mitotic figures were missed during the ground truth annotation, but were then detected by the automatic methods. To further examine this, the false positives from the top two methods, which had notably better performance than the remaining methods, were presented to a panel of two observers for re-evaluation, along with the ground truth mitoses as a control. This panel consisted of one of the observers that initially traversed the slides and one of the observers that participated in the panel that resolved the discordant objects. Separate images of size 200×200 pixels, centered at the false positive detections and the ground truth objects, were extracted and presented in random order to the new panel for re-evaluation, i.e., labelling as mitosis or non-mitosis.

---
[14] The $F_1$-score is equivalent to the Dice overlap coefficient between the set of detections and the set of ground truth objects. The Dice overlap coefficient is used to characterize the inter-observer agreement, which makes it possible to directly compare the detection results to the inter-observer agreement.

The mitotic activity of the tumors is ultimately expressed as the density of mitotic figures, i.e., the number of mitoses per tissue area. To evaluate the performance of the methods for this task, the correlation coefficient between the number of detections and the number of ground truth mitoses per HPF for the subjects in the testing set was computed.

# 5 Results

## 5.1 Mitosis detection

The overall $F_1$-scores along with the precision and recall of the proposed methods are summarized in Table 2 and Figure 4. The top ranking method is IDSIA with an overall $F_1$-score of 0.611.

The individual $F_1$-scores for each subject are summarized in Table 3, along with the average across all subjects. Note that subject #8 from the testing data set has zero annotated ground truth mitotic figures, thus the $F_1$-score is undefined. Instead of the $F_1$-score, for this subject, Table 3 contains the number of false positive detections. According to average $F_1$-score, the top ranking method is again IDSIA with value of 0.445.

Figure 2 gives the PR curves for the IDSIA and DTU methods. The performance of the other methods is also plotted on the same graph for comparison.

Note that these results are based on the original ground truth data and are not influenced by the re-annotation of false positives.

## 5.2 Re-annotation of false positives

The results from the re-annotation experiment are given in Figure 3. The proportion of false positives from the IDSIA method re-annotated as mitotic figures was $p = 0.29$, 95% CI [0.23, 0.36][15]. In other words, out of the 208 false positives from the IDSIA method, 61 were re-annotated as true mitotic figures. The proportion of false positives from the DTU method re-annotated as mitotic figures was $p = 0.16$, 95% CI [0.12, 0.19]. In other words, out of the 397 false positives from the DTU method, 62 were re-annotated as true mitotic figures. This is illustrated in Figure 3A.

The proportion of objects re-annotated as mitotic figures from the entire set of detections produced by the IDSIA method was $p = 0.61$, 95% CI [0.57, 0.65]. This means that out of the 534 detections that this method produced, 326 were re-annotated as true mitotic figures. The corresponding proportion for the DTU method was $p = 0.42$, 95% CI [0.39, 0.46] (293 out of 693 detected objects). For comparison, the proportion of objects re-annotated as mitotic figures from the ground truth dataset was $p = 0.71$, 95% CI [0.67, 0.75] (379 out of 533 objects). This is illustrated in Figure 3B.

## 5.3 Number of mitoses per HPF

The scatter plots for the number of detections and the number of ground truth objects per HPF (averaged for each of the 12 test cases) for the five methods with highest performance according to the overall $F_1$-score is given in Figure 4. The plots for the other methods are omitted for brevity. The best correlation was achieved by the IDSIA method ($r = 0.90$, 95% CI [0.62, 0.96]).

---

[15] CI refers to the 95% confidence interval.

Tables with more detailed results are available for download from the challenge website.

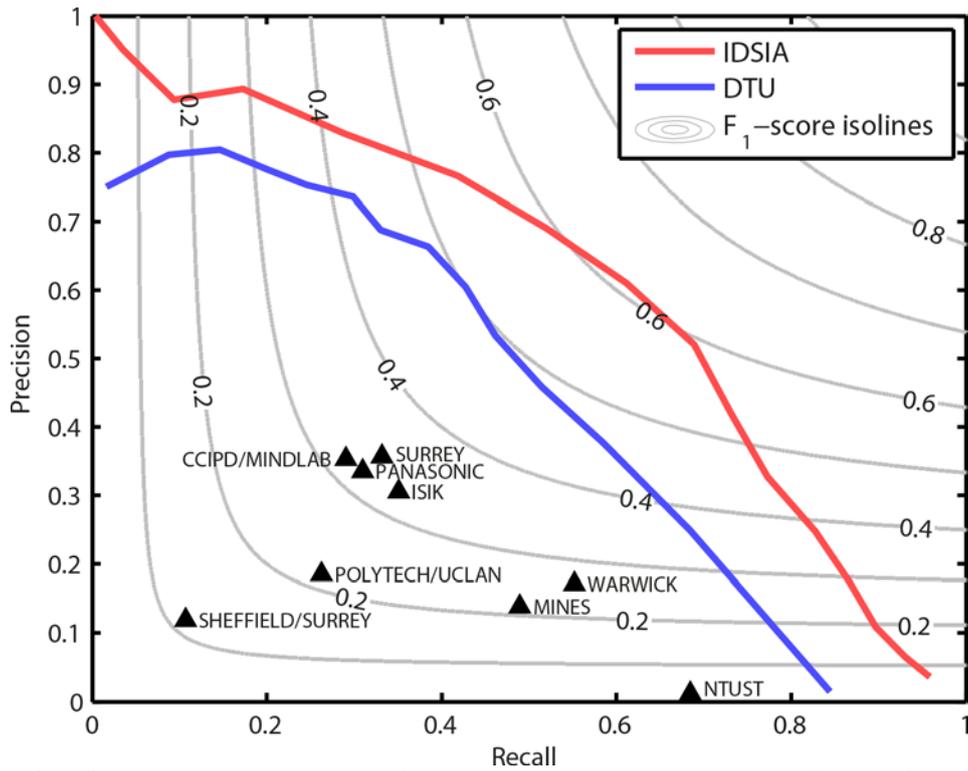

**Figure 2** – Precision-recall curves of the two top ranking methods. The performance of the other methods is plotted for comparison.

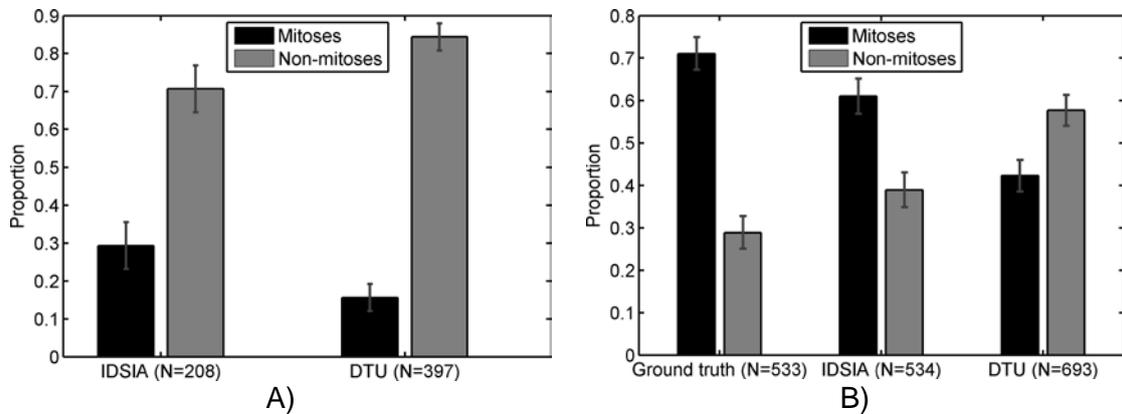

**Figure 3** – Results from the re-annotation of the false positives. A) Proportion of the false positives from the IDSIA and DTU methods that were re-annotated as mitoses and non-mitoses. B) Proportion of the entire set of detections from the IDSIA and DTU methods that were re-annotated as mitoses and non-mitoses. The re-annotation of the ground truth is also given for comparison.

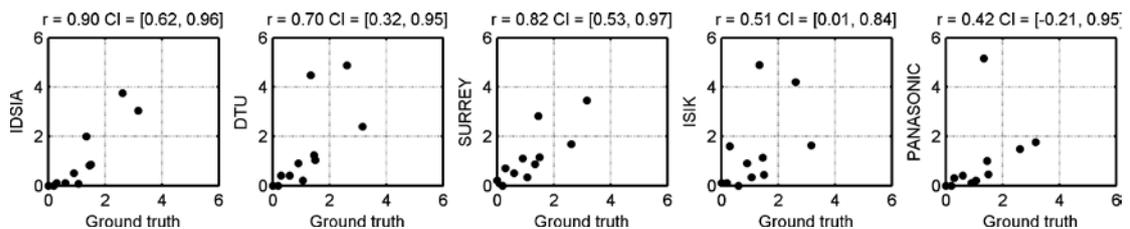

**Figure 4** – Scatter plots for the estimated and ground truth number of mitoses per HPF for the first five methods with highest overall $F_1$-score.

**Table 2** - Overall $F_1$-Scores of the proposed methods. To compute the overall $F_1$-score, all ground truth objects were considered a single data set regardless to which subject they belong.

| Team name | Precision | Recall | F1-Score |
|---|---|---|---|
| **IDSIA** | 0.610 | 0.612 | 0.611 |
| **DTU** | 0.427 | 0.555 | 0.483 |
| **SURREY** | 0.357 | 0.332 | 0.344 |
| **ISIK** | 0.306 | 0.351 | 0.327 |
| **PANASONIC** | 0.336 | 0.310 | 0.322 |
| **CCIPD/MINDLAB** | 0.353 | 0.291 | 0.319 |
| **WARWICK** | 0.171 | 0.552 | 0.261 |
| **POLYTECH/UCLAN** | 0.186 | 0.263 | 0.218 |
| **MINES** | 0.139 | 0.490 | 0.217 |
| **SHEFFIELD/SURREY** | 0.119 | 0.107 | 0.113 |
| **NTUST** | 0.011 | 0.685 | 0.022 |

**Table 3** – Individual (per subject) and average $F_1$-Scores of the proposed methods. To compute the individual $F_1$-scores, every subject was considered a separate dataset.

| Team name | 1 | 2 | 3 | 4 | 5 | 6 | 7 | 8* | 9 | 10 | 11 | Average F1-score |
|---|---|---|---|---|---|---|---|---|---|---|---|---|
| **IDSIA** | .50 | .00 | .66 | .57 | .29 | .73 | .00 | 0 | .46 | .60 | .64 | 0.445 |
| **DTU** | .00 | .11 | .41 | .56 | .20 | .63 | .00 | 0 | .35 | .61 | .68 | 0.352 |
| **WARWICK** | .13 | .34 | .14 | .59 | .40 | .62 | .00 | 3 | .09 | .38 | .34 | 0.302 |
| **ISIK** | .00 | .10 | .37 | .44 | .00 | .47 | .00 | 1 | .16 | .26 | .46 | 0.226 |
| **PANASONIC** | .00 | .00 | .28 | .20 | .40 | .52 | .00 | 0 | .10 | .30 | .33 | 0.213 |
| **CCIPD/MINDLAB** | .00 | .15 | .36 | .38 | .00 | .42 | .00 | 0 | .07 | .38 | .33 | 0.208 |
| **SURREY** | .00 | .10 | .48 | .30 | .00 | .47 | .00 | 2 | .14 | .19 | .38 | 0.205 |
| **MINES** | .00 | .14 | .11 | .40 | .16 | .49 | .00 | 1 | .07 | .32 | .34 | 0.203 |
| **POLYTECH/UCLAN** | .00 | .00 | .07 | .18 | .00 | .50 | .00 | 0 | .04 | .30 | .39 | 0.148 |
| **SHEFFIELD/SURREY** | .03 | .07 | .35 | .15 | .00 | .05 | .00 | 54 | .09 | .08 | .18 | 0.099 |
| **NTUST** | .01 | .03 | .01 | .09 | .03 | .23 | .02 | 29 | .01 | .05 | .21 | 0.068 |

* Subject #8 has no ground truth mitotic figures, thus the $F_1$-score is not defined. In this case, the number of false positives is given in the table. This subject was excluded when computing the average $F_1$-score.

# 6 Discussion

## 6.1 Summary of the proposed methods

The majority of the proposed methods followed a two-step object detection approach. The first step identified candidate objects that were then classified in the second step as mitoses or non-mitoses. Some of the proposed methods prior to the candidate extraction and classification steps performed transformation of the color channels (DTU, PANASONIC, POLYTECH/UCLAN) or staining unmixing (MINES) to obtain a nuclear/hematoxylin channel, thus eliminating eosinophylic structures that can hamper the detection performance. Three of the proposed methods (SURREY, SHEFFIELD/SURREY and WARWICK) perform explicit staining normalization to tackle the problem of staining variability.

The most popular technique for candidate extraction was thresholding of a grayscale image in combination with linear filtering or morphology operators. This technique worked relatively well because of the hyperchromacity of the mitoses – by selecting only the "darkest" nuclei in the images as candidates, a large number of the non-mitoses can be rejected while achieving high sensitivity for the mitosis class. Three of the proposed methods (SURREY, SHEFFIELD/SURREY and WARWICK) used a supervised method to obtain a mitosis likelihood map, which was then thresholded in order to obtain candidate objects.

In the second step, features were computed for segmented mitosis candidates and/or image patches centered at the detected candidate locations. The size of the image patches varied from 63×63 to 128×128 pixels. These patch sizes were selected to be somewhat larger than the expected size of mitotic figures in order to capture contextual information. Spatial pooling was employed by some methods (DTU, POLYTECH/UCLAN) to capture information about structures in different regions of the candidate objects.

A variety of different generic color, texture and shape features, with emphasis on rotational invariance, was employed by the different methods. The CCIPD/MINDLAB method used a combination of classification based on handcrafted features and a feature learning method (convolutional neural networks), and SHEFFIELD/SURREY models the set of observations with manifold learning.

For classification, RBF SVMs (DTU, SURREY, MINES) and random forests (CCIPD/MINDLAB, PANASONIC, POLYTEC/UCLAN) were the most commonly used classifiers, with some methods employing different boosting techniques (ISIK, WARWICK, NTUST). The problem of class imbalance was addressed by subsampling of the dominant negative class or oversampling of the class of mitotic figures.

The IDSIA and NTUST methods did not perform candidate extraction, but instead, evaluated the detector for every pixel location. IDSIA used a very efficient implementation of deep convolutional neural networks to obtain a mitosis probability map for each image, from which mitoses were detected by non-maxima suppression.

### 6.2 Performance of the proposed methods

The best performing method according to all evaluation measures was IDSIA. The overall $F_1$-score of this method was comparable to the inter-observer agreement among pathologists. The DTU method also achieved solid performance according to the overall and average $F_1$-scores. The performance of these two methods was notably better than that of the remaining methods. The analysis of the PR curves (Figure 2) indicated that this is not related to the choice of the operating point of the detectors (the points indicating the performance of the other methods are in the interior of the areas spanned by the curves).

The majority of the false positives produced by the methods with lower performance were dark objects that lack other characteristics of mitotic figures such as protrusions around the edges. On the other hand, many of the mitotic figures that had less intensive staining were not correctly detected. This can be either explained by the fact that the texture and shape features used lack discriminative ability and do not capture these fine structural details or by the fact that the mitotic figures with less intensive staining were underrepresented in the training set. Examples of the most commonly detected and missed mitotic figures are given in Figure 5.

The results from the re-annotation experiment indicate that a large portion of the "false positives" from the IDSIA method can in fact be considered true mitotic figures (Figure 3A). They may have been missed during the ground truth annotation because of the intricacy of the task and the observer variability. In addition, the distribution of the assigned labels during the re-annotation of the original ground truth set and of the set of detections from this method was very similar (Figure 3B).

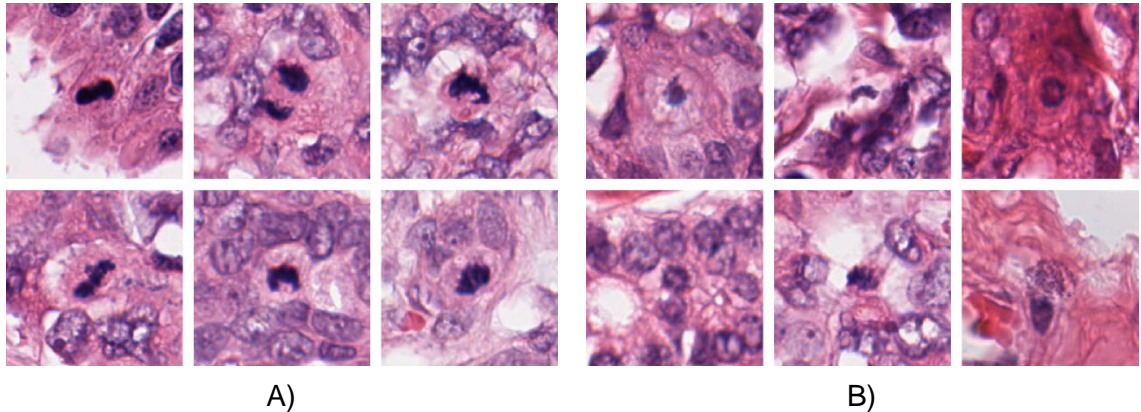

**Figure 5** – Examples of the most commonly detected and missed mitotic figures. A) Mitotic figures that were detected by most (at least ten) of the proposed methods. B) Mitotic figures that were not detected by any of the proposed methods.

Good correlation between the ground truth and estimated number of mitoses per HPF was achieved even for some proposed methods that have lower overall and average $F_1$-scores (Figure 4). This indicates that estimation of the mitotic activity index or the mitotic activity grade might be possible with lower object-level detection performance or even from global image features. This represents an interesting subject for future research. It should be also pointed out that the number of mitoses per HPF does not correspond to the estimated MAI of the outlined regions in the slides because empty HPFs were removed when forming the challenge data set.

Experiments with combining the results from the different methods by majority voting or intersection of the better performing methods did not show improved results over the best individual method. One of the conclusions of the discussion during the challenge workshop was that the variation in the staining appearance is one of the major obstacles for mitosis detection. After the workshops, attempts to improve the IDSIA and DTU methods were made by incorporating explicit staining normalizations that showed promising results. However, due to the preliminary nature of the experiments we chose not to include the results here.

This challenge was focused on evaluating the detection performance; the running times of the algorithms were not evaluated. However, that is a consideration that needs to be made for a potential practical application of the proposed approaches. The best performing method from IDSIA has training time of approximately 3 days per one deep convolutional neural network with an optimized GPU implementation, or approximately 9 days in total for the 3 neural networks that were used in combination. The testing time for one HPF is approximately 31 seconds per variation, which sums to less than 13 min. for all 24 variations (3 neural networks with 8 rotation/mirroring combinations). This implementation does not exploit parallel processing but uses the fast scanning approach detailed in (Giusti et al. 2013). The authors of the DTU method have reported that the training time of the detection algorithm is 1.5 hours on a standard desktop PC with using a naïve Python implementation. The testing of one HPF takes approximately 30 seconds.

Considering that a single slide can consist of thousands of HPFs, further improvement of the testing times for these methods is needed. When evaluating whole slides, a considerable reduction of the running time can potentially be made by using region of interest detectors as a preprocessing step. This can help quickly eliminate areas of the tissue where mitosis detection should not be performed (for example, areas that do not contain tumor tissue).

### 6.3 Performance on individual cases

The worst performance in terms of the $F_1$-score was achieved for cases #1 and #7. They both have very low mitotic activity, and the few ground truth mitoses have atypical appearance (lack of hyperchromacity). The methods that have good overall performance produce a very low number of false positives for these two subjects, suggesting that they would produce a low mitotic activity estimate.

Another case for which very poor performance was achieved is #2. Note that for this case there was large discrepancy between the numbers of objects indicated as mitoses by the two independent observers (Table 1). This indicates that there is an intrinsic difficulty in identifying mitotic figures in this particular case. Case #9 contained many dark nuclei that are not mitotic figures, thus the methods that extensively rely on the staining intensity as a feature had very low performance for this case.

The best performance was achieved for case #6 that had very high mitotic activity and most of the hyperchromatic objects indeed represented mitotic figures.

### 6.4 Feasibility of mitosis counting on whole slide images

Digital slides are still not widely accepted as primary diagnostic modality pending validation studies, with the main concern being the image quality and lack of fine focusing ability (Al-Janabi et al., 2012). In the context of breast cancer histopathology grading, the image quality of whole slide images is principally sufficient for the scoring of nuclear atypia and tubule formation, which together with mitosis counting constitute the commonly used modified Bloom-Richardson grading system. However, the task of mitosis counting is a more delicate one. Detailed examination is required to distinguish mitoses from other mitosis-like objects, which requires the use of fine focusing on the conventional microscope. This feature is missing in whole slide images captured with a single focal plane. Although whole slide imaging scanners that support slide digitization at multiple focal planes are becoming increasingly available, the use of this feature is still rather limited due to the increased scanning time and storage requirements. Taking this into account, there is a possibility of discrepancy between the mitotic activity as estimated by light microscopy and on unifocal whole slide images.

This challenge compared the performance of human experts and computer algorithms on digital slides. It should be emphasized that both the automatic algorithms and the expert observers worked with the same input. The patient and slide selection was done in a way that captures the intra-laboratory variability of the tissue preparation. However, it is likely that the inter-laboratory variability is greater due to different suppliers of reagents, staining protocols and scanners. One possible extension of this challenge is inclusion of data from multiple centers.

At the time when we started the work on annotating the challenge data set, studies that examined the feasibility of mitosis counting on digital slides were lacking. The qualitative impression of pathologists that we interviewed was that the image quality of digital slides imaged at 40× magnification and a single focal plane is sufficient for mitosis counting in most cases, but for some instances it can be difficult to make a firm decision if an object represents mitosis without the possibility to fine tune the focus. We performed a small scale internal validation study and concluded that there is generally a good correlation between the assessment of the mitotic activity index by light microscopy and on whole slide images (Stathonikos et al., 2013). In a recent larger study involving multiple observers and a large number of subjects it was found that the scoring of mitotic activity on whole slide images is as reliable as on conventional glass slides viewed under a microscope (Al-Janabi et al., 2013).

As an alternative to expert annotation, we are currently investigating the use of Phosphohistone H3 (PPH3) labelling of mitotic figures to produce ground truth mitotic figure locations in a more objective manner. These ground truth locations can then be registered to conventionally stained H&E slides. The initial results look promising, but we are still optimizing the procedure.

# 7   Conclusions

In this paper, we summarized the proposed methods and results from a challenge workshop on mitosis detection in breast cancer histopathology images. The challenge data set consisted of 12 subjects for training and 11 for testing, both with more than 500 annotated mitotic figures by multiple observers. In total 14 teams submitted methods for evaluation, 11 of which are described in this paper. The best performing method achieved an accuracy that is in the order of inter-observer variability.

Our intention is for this challenge to be ongoing with incremental extensions of the training and testing data sets. By keeping the challenge website (http://amida13.isi.uu.nl) open for new submissions, we hope to keep a record of the state of the art of mitosis detection in breast cancer histopathology images.

## Acknowledgements

Thomas Walter has received funding from the European Community's Seventh Framework Programme (FP7/2007-2013) under grant agreements number 258068 (Systems Microscopy).

Teofilo E. de Campos would like to thank Neil Lawrence and Andreas Damianou for their helpful suggestions. During this work, he received funds from grants BB/I004769/2 (BBSRC) and EP/F02827X/1 (EPSRC), UK.